\documentclass[conference, 10pt, a4paper]{IEEEtran}

\usepackage{epsfig}
\usepackage{graphicx}
\usepackage{amsmath}
\usepackage{amssymb}
\usepackage[utf8]{inputenc}
\usepackage{subfigure}
\usepackage{ algorithmic }
\usepackage{multirow}
\usepackage{url}
\newcommand{\nbusers}{83}
\newcommand{\nbgenuinesamples}{5185}
\newcommand{\nbimposedsamples}{5439}
\newcommand{\nbimpostorsamples}{5754}

\begin{document}

%%%%%%%%% TITLE
\title{Web-Based Benchmark for Keystroke Dynamics Biometric
Systems: A Statistical Analysis}

\author{Romain Giot \and  Mohamad El-Abed \and Christophe Rosenberger\\
Université de Caen Basse-Normandie, UMR 6072 GREYC, F-14032 Caen, France\\
ENSICAEN, UMR 6072 GREYC, F-14050 Caen, France\\
CNRS, UMR 6072 GREYC, F-14032 Caen, France\\
\url{firstname.lastname@ensicaen.fr}}

\author{\IEEEauthorblockN{Romain Giot and Mohamad El-Abed and Christophe Rosenberger}
\IEEEauthorblockA{Université de Caen, UMR 6072 GREYC\\
ENSICAEN, UMR 6072 GREYC\\
CNRS, UMR 6072 GREYC\\
Email:
\{romain.giot,mohamad.elabed,christophe.rosenberger\}@ensicaen.fr}}

\maketitle
% \thispagestyle{empty}

%%%%%%%%% ABSTRACT
\begin{abstract}
Most keystroke dynamics studies have been evaluated using a specific kind of dataset in which users
type an imposed login and password. Moreover, these studies are
optimistics since
most of them use different acquisition protocols, private datasets, controlled
environment, etc. In order to enhance the accuracy of keystroke dynamics'
performance, the main contribution of this paper is twofold. First, we provide a
new kind of dataset in which users have typed both an imposed and a chosen
pairs of logins and passwords. In addition, the keystroke dynamics samples are
collected in a web-based uncontrolled environment (OS, keyboards, browser,
etc.). Such kind of dataset is important since it provides us more realistic
results of keystroke dynamics' performance in comparison to the
literature (controlled environment, etc.). Second, we present a
statistical analysis of well known assertions such as the
relationship between performance and password size, impact of
fusion schemes on system overall performance, and others such as
the relationship between performance and entropy. We put into
obviousness in this paper some new results on 
keystroke dynamics in realistic conditions. 

%enhance the optimistic resutls of keystroke dynamics performance 
%
%In order to enhance the accuracy of keystroke dynamics performance results, 
%the key contribution o
%
%
%In this study, we have collected (and used) another kind of dataset in which users have typed 
%an imposed login and password, and a chosen login and password. In order to enhance the accuracy 
%of the presented results, the samples have been collected in a web-based uncontrolled environment 
%(OS, keyboards, browser, etc.). Such kind of acquisition is important since it provide us more 
%realistic results of keystroke dynamics performance in comparison to the literature (controlled 
%environment, private datasets, etc.). We put into obviousness in this paper some new results on 
%keystroke dynamics in realistic conditions. Experimental results show that the performance of existing methods
%in the state of the art are over-estimated in real conditions. 
%We have statistically verified well known assertions such as the relationship between
%performance and password size, impact of fusion schemes on system overall performance, and others such as the relationship 
%between performance and entropy.

\end{abstract}

%%%%%%%%% BODY TEXT
\section{Introduction}

%\todo[inline]{il faudrait encore motiver le papier dans l'intro
%Ensuite, tu
%pars de la base et définit les expériences. Je partirai des questions en se
%positionnant du côté opérationnel; je veux mettre en place un service web avec
%une authentification par dynamique de frappe au clavier (pense à l'application
%TES). Quelles sont tes questions ? qu'est ce qui existe ? quelles sont les
%limites des études existantes ? ensuite, tu définis une approche de résolution :
%benchmark, méthode statistique d'analyse, complexité d'un mot de passe,
%fusion... et enfin tu présentes des résultats expérimentaux reprenant la réponse
%aux questions
%en mettant tout dans experiment, tu ne mets pas en valeur le
%travail.}
%
Keystroke dynamics allows users to be recognized based on their way of typing on a keyboard.
This is a behavioral modality which has been first experimented in the
eighties~\cite{gaines1980}.
It is always an interesting subject of research, as it is a low cost two factors authentication approach.
Most consequent keystroke dynamics studies have been evaluated on datasets where users
typed the same fixed string~\cite{killourhy2009cad,giot2009benchmark,balagani2011on},
while very few of them used different strings for each
users~\cite{cho2000wbk,Hosseinzadeh2008GMM}.

For this study, we want to be placed in the following context.
We want to use a web-based application with an authentication system based on static keystroke dynamics.
Some studies have already been done on web-based keystroke dynamics~\cite{Revett2005DKD, cho2000wbk, Kofi2005AKD,jiang2007ksl},
but none of them provided the used dataset and their experimental
protocols were really differents.
Some worked with individual passwords~\cite{jiang2007ksl}, while other used the same strings for each user~\cite{Revett2005DKD}.
In our work, we statistically analyze the behavior of these two approaches.

There is a strong need of
a large dataset.
%(both in terms of number of users and sessions). 
We provide a new dataset where users typed both an imposed pair of login and password,
a chosen login (their usual one) and password (one chosen by themselves for the experiment).
The aim of the dataset is to show the viability of using personal
identifiers (\emph{i.e.}, chosen login and password) in native web browser
(\emph{i.e.}, using no plug-in or extension of the web browser),
because the most recent applications are web-based ones, and systems usually use different logins and passwords for each user.
The contribution of this paper is to present this new dataset that 
is publicly available\footnote{\url{http://www.epaymentbiometrics.ensicaen.fr/}} for testing algorithms in an operational context and experiment keystroke dynamics on this dataset.
It is the sole public dataset which satisfy these properties.
Additionally, we analyze information provided in this dataset to
answer operational questions such as those presented in section
III, part B.
The paper is organized as follows.
Section~\ref{sec_dataset} presents existing public datasets and the dataset
built for the experiment.
Section~\ref{sec_experiment} presents the various experiments.
Section~\ref{sec_results} presents their results.
Section~\ref{sec_conclusion} concludes this paper and gives some perspectives.

\section{Public Keystroke Dynamics Datasets}
  \label{sec_dataset}
%In the next section, we first make a state of the art concerning datasets in keystroke dynamics.
%\subsection{Existing Public Datasets}
%Before experimenting on keystroke dynamics, it is necessary to have a dataset.
In most studies, researchers use their own dataset which, most of the time, suffers of lack of number of users and sessions.
Some keystroke dynamics databases are publicly available in the literature, but
none of them provides different login and password for each user.
In~\cite{giot2009benchmark}, several users have typed the passphrase ``greyc
laboratory'' on two different keyboards on the same computer during several
sessions.
100 users have provided at least 60 samples each on 5 different
sessions spaced of one week (most of the time).
This database contains the most number of users, but, the number of samples and
sessions may be too small to track variability through time.
In~\cite{killourhy2009cad}, several users have typed the password ``.tie5Roanl'' on a
single computer during several sessions.
51 users have provided 400 samples each on 8 different sessions spaced
of, at least, one day.
This database contains a huge number of samples, but the time interval may be
too small to track variability on a long period.
These two databases are the only ones containing enough samples and users to
give statistically significant results.
Sadly, they mainly have been used by their own creators, and not by the
community. 
Table~\ref{tab_databases} summarizes this information.
Even if these two databases are interesting, they do not really fit
requirements for realistic studies:
\begin{enumerate}
  \item 
%(a) 
We want different logins and passwords per user. This is the most
realistic scenario for keystroke dynamics.
%although very few keystroke dynamics
%studies work with such a realistic dataset~\cite{Hosseinzadeh2008GMM, Revett2007mlaa}. 
Real users use different logins 
and passwords. 
%The other public databases only fit systems where each user types a
%predefined text (which is not secret).
  \item 
%(b) The capture must be done on different computers and keyboards.
%In~\cite{giot2009benchmark}, two different keyboards (which allows a variability
%due to the shape of the keyboard) are used, but, in
%both~\cite{giot2009benchmark, killourhy2009cad} the same computer is used for all the samples.
It is interesting to have different computers and keyboards to grow the
variability of the samples (due to shape of keyboard, responsiveness of the computer, precision
of the timer, \ldots).
   \item 
%(c) 
The captures must have been done in a web browser (because nowadays, most of
modern applications are available as web-based applications, and not desktop
applications.
Collecting samples from different browsers allows to track more variability due
the browser itself (several browsers exist on all the operating systems).
\end{enumerate}

We have created a web-based application which allows us to capture keystroke
dynamics during several sessions.
We think that with this dataset, researchers will have an interesting dataset
providing a lot of variabilities due to the different factors presented before.
The results would not be over optimistic as it may be the case with actual ones.
The next section presents the experiment.

\section{Experiment}
  \label{sec_experiment}
%As there is no public dataset fitting our requirements, we have created  our own dataset.
%After presenting the dataset, we present the distance computation's method we used,
%the way of statistically validates the experiments, and the various studies.

\subsection{Proposed Dataset}
\subsubsection{Acquisition Protocol}
Each week, we have sent an email to %the mailing list of 
the students of our
school of engineering and some
colleagues of our lab.
It asks them to realize the session capture of the week.
The first one explains the aim of their participation.
During the first session, each user has to choose its own login (we asked them
to use the login of their school account, but they have not all
respected that), and password.
We expect them to type their login as they are used to.
So, each user chooses when he/she wants to do the session without any obligation
or pressure.
Participants have not been rewarded.
A session is composed of three different steps.
Each step consists in typing several times a pair of login and password.
The user interface presents two input fields: one for the login, and one for
the password.
No typing correction is allowed: if a user presses backspace, the input field
is cleared, and the user must type its text from scratch.
The password and login the user have to type are displayed near the form and are displayed
in a pop-up box at the start of each step\footnote{No screenshots for 
lack of space}.

A progression bar is shown at the bottom of the screen.
It indicates how many inputs are yet needed to complete the session.
As the interface is displayed in a web browser, it is written with Javascript, html
and css.
Most studies of keystroke dynamics which work on a web browser are written in
Java~\cite{Revett2005DKD, cho2000wbk, Kofi2005AKD}. 
We have not chosen this language because it imposes the user to install a Java
plugin for its browser.
%(91\% of the browsers used for the experiment have such
%plugin whereas 100\% have activated Javascript).
For each key event (key press and key release), the timing information is
captured through the $timeStamp$ value of the event\footnote{\url{https://developer.mozilla.org/En/DOM/Event.timeStamp}}.
We do not track timing information of the key having a code inferior to 48
(except tab, shift, space, ctrl, altgr, and keycode 0 which seems to be present
for
some punctuated keys) as all as right and left Windows key and keys from F1 to
F12.
The three acquisition steps are the following ones:
\begin{itemize}
\item \emph{Step 1}. 
%(1) 
Ten inputs of a pair of imposed login and password.
This allows us to capture exactly the same thing for all the users as
in~\cite{giot2009benchmark, killourhy2009cad}.
% except we capture both login and
%password instead of a single password.
\item \emph{Step 2}. 
%(2) 
Ten inputs of the chosen pair of login and password of the
user.
This simulates the authentication of the user on a system as in~\cite{Hosseinzadeh2008GMM, Revett2007mlaa}.
\item \emph{Step 3}.
%(3) 
For two other users, five inputs of the selected pair of login and
password.
This allows us to capture ten impostor samples belonging to two other users.
\end{itemize}

\subsubsection{Presentation of the Obtained Dataset}
As the participation was only based on the goodwill of the users, very few of
them participated to the study or to each
session.
That is why only $\nbusers$ users have participated to the study, whereas the emails were
sent to more than $300$ students.
Sessions were not always done completely.
Users have provided a total of $\nbgenuinesamples$ genuine samples (pair of 
login, password typed by its owner), $\nbimpostorsamples$ impostor samples
(pair of login, password typed by a user different of its owner), and
$\nbimposedsamples$ imposed samples (pair of imposed login and password).
%Of course, the dataset will be made publicly available as soon as possible in this website: \url{http://www.epaymentbiometrics.ensicaen.fr/}.
Most participants are between 20 and 24 years old (mainly students
in computer science, chemistry and electronics).
Most users are males, which can be problematic to generalize
results if males and
females have different typing behaviors~\cite{giot2011anew}.
Most users have more than 20 impostor samples which allows to
obtain good information on False Match Rate.
The number of genuine and imposed samples per user is not equally set, there are
several users who have provided less than 40 genuine samples.
It is difficult to obtain a large keystroke dynamics
database with enough quantity of samples per user (which may explain why there
are so few publicly available keystroke dynamics databases, and why, most of the
time the number of users is relatively small).

\begin{table}[!tb]
\centering
\caption{Summary of the keystroke dynamics databases.
Prop. refers to the proposed dataset.
Same means each user types the same password (so impostor are used to type the
same password than the user), while different means each user types a different
password (and impostors are not used to type it).}
\label{tab_databases}
\begin{tabular}{|c|c|p{2cm}|c|c|}\hline
 &
 \multicolumn{2}{|c}{Size} &
 \multicolumn{2}{|c|}{Login/password}\\\hline
\textbf{Study} &
\textbf{\# users} &
\textbf{\# samples} &
\textbf{same} &
\textbf{different} \\ \hline

\cite{giot2009benchmark} & $100$ & $60000$ & $\checkmark$ & \\ \hline
\cite{killourhy2009cad} & $51$ & $20400$ & $\checkmark$ & \\ \hline
Prop. & \nbusers & $\nbgenuinesamples + \nbimpostorsamples / \nbimposedsamples $ &
$\checkmark$ & $\checkmark$ \\\hline
\end{tabular}
\end{table}

Although the obtained dataset is not the largest in terms of number
of users involved, it is the only public keystroke dynamics
providing different logins and passwords per users.
Thus, it is the more realistic one.

\subsection{Experimental Protocol}
We want to answer to the following questions:
\begin{enumerate}
 \item 
Does keystroke dynamics' performance behaves similarly on a dataset
built with imposed strings, against a dataset built with strings chosen by users
themselves? 
This question is interesting because all public
datasets do use imposed strings.
 \item 
Which approach (individual or global threshold) gives better results in terms of 
performance?
This question is interesting, because both approaches are used in
the literature and avoid an easy performance comparison.
 \item 
Which features must be used in a score fusion system, in order to improve
results?
 \item 
Are password length, entropy and complexity correlated
with the recognition performance?
This question is interesting, because it has not been studied in
the literature (probably because all the passwords are identical).
It can give information of how the password must or must not be
chosen by the user, in order to strengthen the system.

\end{enumerate}

%To our knowledge, first and last questions have never been explored before. 

We have run several experiments using the different subsets (chosen/imposed)
to analyze the performance of keystroke dynamics authentication methods.
The Equal Error Rate (EER) is individually computed for each user
(\emph{i.e.}, EER is computed with the
comparison scores of its test samples against its model and real impostors' test samples against its model), and,
its averaged value (among all users) is presented under $EER_i$.
$EER_g$ presents the EER with the same threshold for all the users
(EER is computed with a global intra-scores and inter-scores set).
These are two common ways of presenting keystroke dynamics results, but no study analyzed the performance difference between the two approaches.
Authentication test is done with only one capture, we do not give another chance
if it fails (several tries is a another common way of presenting
results~\cite{Hosseinzadeh2008GMM, killourhy2010kbn}).
%Most keystroke dynamics studies only work with the username of the user, or a
%passphrase. 
%We have done such experiment in section~\ref{sec_simple}.
%As we provide the password and the username of the user, we have done
%experiments which use a score fusion of these two different strings to
%authenticate users in section~\ref{sec_score_fusion}.
Training is done with 20 samples (two sessions), and testing is done with the
remaining samples. We only keep users having at least 20 testing samples (at least two sessions per user).
So, we work with users having used the system during at least 4 sessions.
As users may use different keys for typing their login or password, the
number of pressed characters may be different.
%This may held to a certain number of Failure To Acquire Rate if not enough
%samples are available to compute the model.

\subsubsection{Distance Computation}\label{sec_distance_coputation}
In this paper, we have tested only one score computing method.
It is based on a Gaussian distribution assumption of the
features~\cite{hocquet2006eou}.
Each user provides $N$ samples to build its template.
A sample $\mathbf{x}$ is a $n$-dimension vector.
The template $\boldsymbol{\theta=}(\boldsymbol{\mu},\boldsymbol{\sigma})$ is composed of the
mean vector $\boldsymbol{\mu}$ and the standard
deviation vector $\boldsymbol{\sigma}$ among these features.
The distance between sample $\mathbf{x}$ and template $\boldsymbol{\theta}$ is computed using the following formulation:
\begin{equation}
%$
d(\mathbf{x}, \boldsymbol{\theta}) = 
1 - \frac{1}{n} 
\sum_i^n exp\left( -\frac{|x_i - \mu_i|}{\sigma_i} \right)
%$.
\end{equation}
If a query is not of the same size (different combinations of keys, or use of
the mouse to select and erase text, can be the
reason of this difference) of the template, we return a distance of 1 (the
worst score, 0 being the best).

\subsubsection{Statistical Validation}
In order to verify the various
answers, we use the 
Kruskal-Wallis (KW) test \cite{stat2003higgins}. 
It is a non-parametric (distribution free) test, 
which is used to decide whether $K$ independent samples are from the 
same population. 
More generally speaking, it is used to test two hypothesis: 
the null hypothesis ($H_0:~\mu_{1}~=~\mu_{2}~= ...=~\mu_{k}$) 
assumes that samples have been generated
from the same population (\emph{i.e.}, equal population means) against the alternative 
hypothesis ($H_1:~\mu_{i}~\neq~\mu_{j}$) which assumes that there is a statistically significant 
difference between at least two of the subgroups. The decision criterion is then derived 
from the estimated $p-value$ as depicted in equation \ref{Eq:krusk_p_value}.

\begin{equation}
\label{Eq:krusk_p_value}
\left\{
\begin{array}{ll}
p-value ~\ge~0.05~ & ~accept~H_{0}\\
otherwise~&~ reject~H_{0}
\end{array}
\right.
\end{equation}

\subsubsection{Simple Feature Authentication}\label{sec_simple}
We test several features: ``rp'' (latency between the release of a key, and
the pressure of next one), ``rr'' (latency between the release of two successive
keys), ``pp'' (latency between the pressure of two successive keys) and ``pr'' (duration of pressure of one key) for both 
login and password individually, and for each kind of datasets (imposed 
login/password and chosen login/password).
This gives us $4*2*2=16$\footnote{number of features * login or password
* imposed or chosen} different experiments.

\subsubsection{Score Fusion}\label{sec_score_fusion}
\begin{figure}
  \centering
  \includegraphics[width=0.8\linewidth]{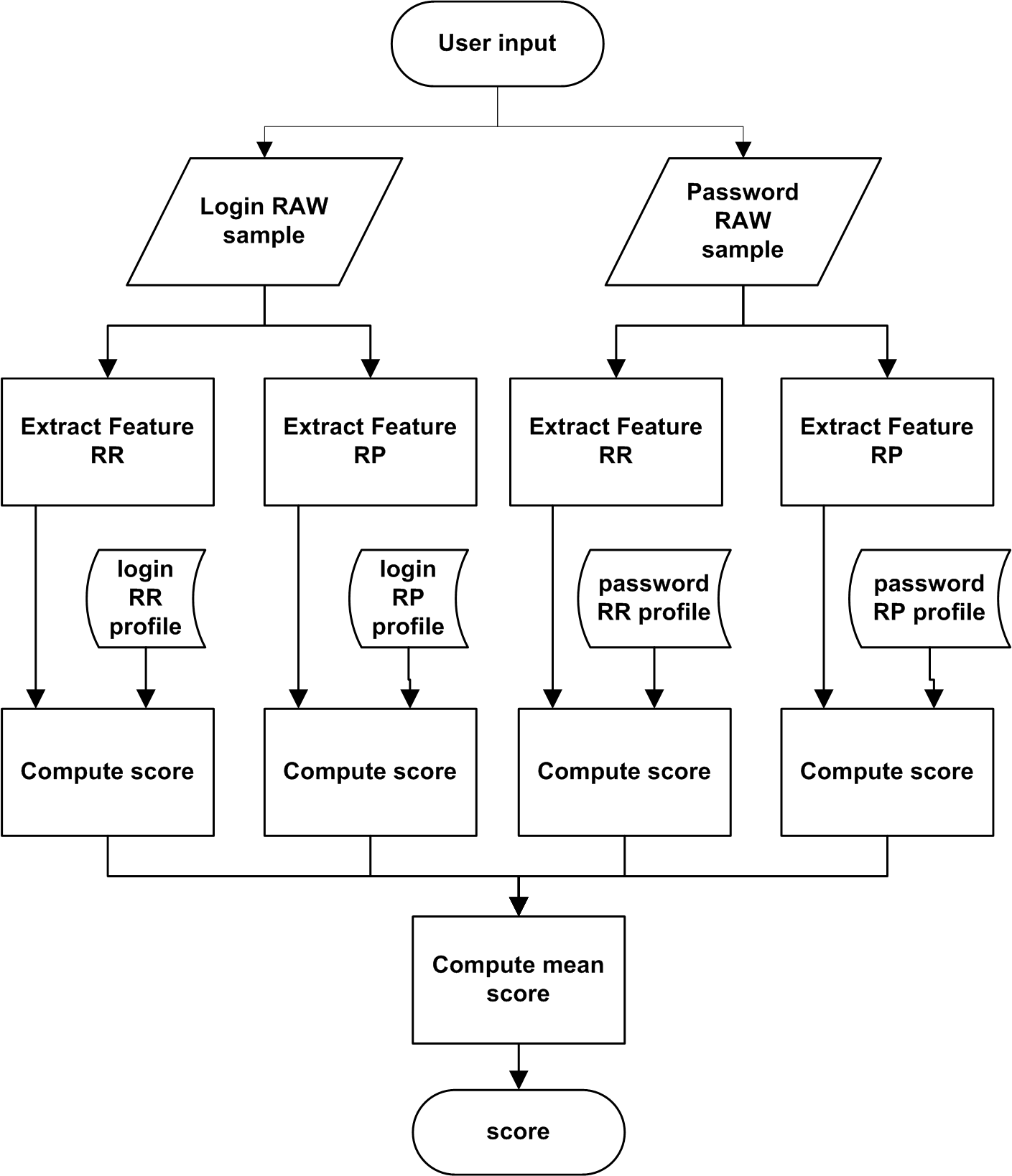}
  \caption{Score fusion scheme using only RR and RP features.}
  \label{fig_score_fusion}
\end{figure}

Although feature fusion is often used in keystroke
dynamics~\cite{giot2009svm,balagani2011on,killourhy2009cad}, we have chosen to use a
score fusion system~\cite{shen2001statistical, hocquet2006eou}.
A user sample is composed of several sub-samples (one per kind of extracted
features):
$\mathbf{x}=(\mathbf{x}_{rr}^l,\mathbf{x}_{rp}^l,\mathbf{x}_{pr}^l,\mathbf{x}_{pp}^l,
\mathbf{x}_{rr}^p,\mathbf{x}_{rp}^p,\mathbf{x}_{pr}^p,\mathbf{x}_{pp}^p)$
(superscript $l$ and $p$ respectively represent login and
password).
A template is built for each kind of sample extracted features: 
$\boldsymbol{\theta}=(\boldsymbol{\theta}_{rr}^l,\boldsymbol{\theta}_{rp}^l,\boldsymbol{\theta}_{pr}^l,\boldsymbol{\theta}_{pp}^l,
\boldsymbol{\theta}_{rr}^p,\boldsymbol{\theta}_{rp}^p,\boldsymbol{\theta}_{pr}^p,\boldsymbol{\theta}_{pp}^p)$.
In this case, the same keystroke dynamics method is used for each extracted
features.
The final score is the mean (without score normalization) of each of these scores (one for each selected feature), so the fusion rule is:
%\begin{equation}
$
s_f = \frac{1}{m}\sum_{i}^m s_i
$
%\end{equation}
\noindent with $s_f$ the new fused score, and $s_i$ the comparison score of
system $i$ (using features of type $i$), when using $m$ different systems.
%Simple fusion functions work well, and, it was not the purpose of this paper to
%chose better fusion functions (although we expect to obtain better results by
%using a more weighted score fusion function).
As an illustration, figure~\ref{fig_score_fusion} presents the score fusion architecture when pp and
rp times from login and password are used.

%Other fusion function for keystroke dynamics systems can be found
%here~\cite{giot2010}.

\subsubsection{Study of the performance depending on password size and complexity}
\label{sec_correlation}
For this experiment, we use the score fusion of all the features of the password.
We would like to verify if keystroke dynamics' performance 
depends on the complexity, the size, or the entropy of the password. Towards this 
goal, we use again the KW test. For the complexity computation, 
we have used an existing algorithm which is often used in web
applications
as depicted in figure~\ref{fig_compute_complexity}.
The entropy quantifies the expected value of the information contained in the
password $\mathbf{p}$.
The password contains $P$ unique characters $\{c_1,\cdots,c_P\}$.
$p(c_i)$ is the probability of appearance of the character $c_i$ in the password
$\mathbf{P}$.
Entropy is computed as following:
\begin{equation}
  H(\mathbf{P}) = - \sum_{i=1}^P p(c_i)\log_2(c_i)
\end{equation}

%To confirm or not the relationship between performance and password complexity (respectively password size), we use the Pearson Correlation Coefficient between the individual EER of each user and the password complexity (respectively size).
%The relationship is confirmed if the coefficient is bigger than 0.5 and the p-value lower than 0.05.
%The correlation factors ($\rho_{complexity} = -0.28$, $\rho_{length} = -0.32$) which is not enough important to assert a correlation between password complexity (respectively password size) and performance.
%Figure~\ref{fig_compute_complexity} presents the way of computing the complexity
%of a string.

\begin{figure}
 \centering
\footnotesize
\begin{algorithmic}
\REQUIRE $PASSWORD$

\STATE $SIZE$ $\leftarrow$ number of char of $PASSWORD$
\STATE $LOW$, $UPP$, $NUM$, $OTH$ $\leftarrow$ (0, 0, 0, 0)

\FOR{$i=1$ \TO $SIZE$}
  \STATE $CHAR$ $\leftarrow$ $PASSWORD[i]$
  \IF{$CHAR \geq$ 'a' and $CHAR \leq$ 'z'}
	\STATE $SCORE \leftarrow SCORE + 1$
	\STATE $LOW \leftarrow 1$
  \ELSIF{$CHAR \geq$ 'A' and $CHAR \leq$ 'Z'}
	\STATE $SCORE \leftarrow SCORE + 2$
	\STATE $UPP \leftarrow 2$
  \ELSIF{$CHAR \geq$ '0' and $CHAR \leq$ '9'}
	\STATE $SCORE \leftarrow SCORE + 3$
	\STATE $NUM \leftarrow 3$
  \ELSE
	\STATE $SCORE \leftarrow SCORE + 5$
	\STATE $OTH \leftarrow 5$
  \ENDIF

\ENDFOR
  \STATE $COEFF \leftarrow SCORE/SIZE$
  \STATE $DIVERSITY \leftarrow LOW + UPP + NUM + OTH$
  \RETURN $COEFF* DIVERSITY * SIZE$
\end{algorithmic}
\caption{Compute the complexity of a password (in terme of
password security and not typing difficulty).}
\label{fig_compute_complexity}
\end{figure}

\section{Results}
  \label{sec_results}
In this section, we present the results of the experiments previously
presented.
Even if the number of users in the dataset is quite important, only 48 of them
provided enough samples to be used in the experiments (note that most of keystroke dynamics studies even use fewer individuals).

\subsection{Simple Feature Authentication}
Table~\ref{tab_result_simple} presents the results of the simple feature
authentication experiments.
Using the KW test, we find that there is no significant difference ($p-value=0.68$) 
of performance between the chosen and the imposed datasets. This was a surprising 
result since users are more likely to type their own \textit{login} 
and \textit{password} 
than an imposed ones. 
We may obtain these results because, even if users have chosen their password,
it is not their real password they type several times per day.
%A further investigation of this question should be done 
%when having a larger dataset. 
Using the chosen and imposed datasets, we found 
that the performance of individual approach outperformed 
($p-value<<0.05$) the global approach. 
%For the performance of using the 
%\textit{login} or \textit{password} information, we found that: 
%\begin{itemize}
%    \item 
%The use of the \textit{password} outperformed ($p-value=0.02768$) the 
%\textit{login} using the chosen dataset. This 
%result was attended since the users are used to type their own passwords in an easy 
%manner (hence, legitimate users typing is much more stable than an impostor one).  
%
%    \item 

Using 
both datasets, 
we found that the \textit{login} outperformed 
($p-values$ are below to $0.05$, respectively) the \textit{password} information. 
This result was also attended since the used logins are much more easier than passwords 
(hence, users' way of typing the imposed logins would be much more stable than typing 
the imposed passwords).
%\end{itemize}

%From these results, we can conclude that the use of \textit{login} information outperformed 
%the \textit{password} one. 
A study of the fusion of both information is given in the next 
section.

%We can also point out that this two results are contradictory, but, they can be
%also be explained because in the imposed dataset passwords are usually longer than
%logins while in the imposed dataset login is longer than password.
%Dependency between length and performance is verified in
%section~\ref{sec_correlation}.

\begin{table}[!tb]
\centering
\caption{Authentication results, for different extracted features for each kind
of text.
The best result of each line is in bold.
The best result of each column is underlined.}
\label{tab_result_simple}

\begin{tabular}{|l|l||r|r||r|r|}
\hline
 &  & \multicolumn{ 2}{c||}{\textbf{Chosen dataset}} & \multicolumn{ 2}{c|}{\textbf{Imposed dataset}} \\ \hline
\textbf{Type} & \textbf{Field} & \multicolumn{1}{l|}{\textbf{EERi}} &
\multicolumn{1}{l||}{\textbf{EERg}} & \multicolumn{1}{l|}{\textbf{EERi}} & \multicolumn{1}{l|}{\textbf{EERg}} \\ \hline
\textit{pr} & \textit{login} & 26.50\% & 28.81\% & \textbf{19.79\%} & 21.90\% \\ \hline
\textit{rp} & \textit{login} & 21.25\% & 25.91\% & \textbf{14.84\%} & 20.91\% \\ \hline
\textit{rr} & \textit{login} & 18.00\% & 24.01\% & \underline{\textbf{10.00\%}} &
\underline{15.21\%} \\ \hline
\textit{pp} & \textit{login} & 19.27\% & 24.48\% & \textbf{11.86\%} & 18.63\% \\
\hline \hline
\textit{pr} & \textit{pwd} & \textbf{22.21\%} & 25.30\% & 23.21\% & 27.08\% \\ \hline
\textit{rp} & \textit{pwd} & \textbf{18.51\%} & 21.56\% & 26.63\% & 30.54\% \\ \hline
\textit{rr} & \textit{pwd} & \textbf{16.95\%} & \underline{19.90\%} & 22.17\% & 27.00\% \\ \hline
\textit{pp} & \textit{pwd} & \underline{\textbf{16.45\%}} & 20.64\% & 24.02\% & 29.00\% \\ \hline

\multicolumn{1}{c|}{}    & \textbf{Mean}& \textbf{19.89\%} & 22.57\% & 19.65\% & 23.78\%\\
\cline{2-6}
\end{tabular}

\end{table}

\subsection{Score Fusion}
Table~\ref{tab_result_fusion} presents the performance of the score fusion when
using different features on the chosen dataset.
We can see that we can improve the performance by fusing the comparison scores
of the right extracted features.
%We can see that using all the features (even if they are redundant) improves the performance.
%This generalizes results obtained on a single fixed text~\cite{balagani2011on}.

In order to see which feature (or combination of features) gives the best performance 
result, we use the KW test over the seven sets which combine login with password: the EER values related to the use of 
``pr'', ``rr'', ``pp'', ``pr'' \& ``rr'', ``pr'' \& ``pp'', ``rr'' \& ``pp'', and 
``pr'' \& ``rr'' \& ``pp'' informations, respectively. We found that the worst 
result (with $p-value$ below to $0.05$) is obtained by using the ``pr'' information (the duration of the press of a key which is the most often used feature). 
We have not selected ``rp'' which is the worst feature in
Table~\ref{tab_result_simple}.
The use of all the features (``pr'' \& ``rr'' \& ``pp'') outperformed 
($p-values$ below to $0.05$) 
the use of ``pr'' and ``pp'' informations, while there were no significant 
performance difference between all the features and ``rr'', ``pr'' \& ``rr'', 
``pr'' \& ``pp'', ``rr'' \& ``pp'' informations. We conclude, that the use 
of all the features (even if they are redundant) may improve the performance. 
This generalizes results obtained on a single fixed text~\cite{balagani2011on}.

\begin{table}[!tb]
\centering
\caption{Authentication results, when using various feature fusion.
The best result of each line is in bold. The best overall result is underlined.}
\label{tab_result_fusion}
\begin{tabular}{|c|c|c|c|c|c|c|c|}  \hline
 \multicolumn{3}{|c}{\textbf{Login}} & 
 \multicolumn{3}{|c|}{\textbf{Password}} &
 \multirow{2}{*}{\textbf{$\text{EER}_\text{i}$}} &
 \multirow{2}{*}{\textbf{$\text{EER}_\text{g}$}}
									     						\\ 
\cline{1-6}
\textbf{pr}  &  \textbf{rr} & \textbf{pp}  & \textbf{pr}  & \textbf{rr}  & \textbf{pp}  &        &       \\\hline

\multicolumn{8}{|c|}{Login only}\\\hline

$\checkmark$&               & $\checkmark$ &              &              &              & \textbf{15.99\%} &	22.42\% \\ \hline	
$\checkmark$& \checkmark    & $\checkmark$ &              &              &              & \textbf{14.36\%} &	20.72\% \\ \hline	
            & \checkmark    & $\checkmark$ &              &              &              & \textbf{16.57\%} &	23.07\% \\ \hline	\hline

\multicolumn{8}{|c|}{Password only}\\\hline
&&&$\checkmark$&               & $\checkmark$ &               \textbf{14.24\%} &	17.75\% \\ \hline	
&&&$\checkmark$& \checkmark    & $\checkmark$ &               \textbf{12.52\%} &	16.74\% \\ \hline	
&&&            & \checkmark    & $\checkmark$ &              \textbf{15.36\%} & 19.04\% \\ \hline	\hline

\multicolumn{8}{|c|}{Login and password}\\\hline
$\checkmark$ &              &              & $\checkmark$ &              &              & \textbf{18.92\%} & 23.51\%  \\ \hline
             & $\checkmark$ &              &              & $\checkmark$ &              & \textbf{12.37\%} &   15.63\%   \\ \hline
             &              & $\checkmark$ &              &              & $\checkmark$ & \textbf{11.45\%} & 16.15\%  \\ \hline
%$\checkmark$ & $\checkmark$ &              &              &              &              & \textbf{16.19\%} & 22.04\%  \\ \hline
%             &              &              & $\checkmark$ & $\checkmark$ &              & \textbf{13.68\%} & 17.58\%  \\ \hline
$\checkmark$ & $\checkmark$ &              & $\checkmark$ & $\checkmark$ &              & \textbf{10.25\%} & 15.85\%  \\ \hline
$\checkmark$ &  &  $\checkmark$            & $\checkmark$ &  &      $\checkmark$        & \textbf{9.4\%} & 19.96\%  \\ \hline
             & $\checkmark$ & $\checkmark$ &              & $\checkmark$ & $\checkmark$ & \textbf{10.71\%} & 14.95\%  \\ \hline
$\checkmark$ & $\checkmark$ & $\checkmark$ & $\checkmark$ & $\checkmark$ & $\checkmark$ & \underline{\textbf{08.87\%}} & \underline{14.08\%}  \\ \hline

        \multicolumn{4}{c|}{}
& \multicolumn{2}{c|}{\textbf{Mean}}& \textbf{13.15\%} & 18.45\%  \\ 
\cline{5-8}
\end{tabular}%                            #
\end{table}

\subsection{Study of the performance
depending on password size, entropy and complexity}
\label{sec_correlation}
Using the KW test, we find that the size ($p-value=0.019$) and the entropy 
($p-value=0.0062$) of the used passwords have a 
significant impact on system performance, while there was no impact 
($p-value=0.12$) according 
to the complexity algorithm. More generally speaking, the average EER 
value is increased 
from $10.03\%$ (for users having more than $8$ characters) to $15.85\%$ (for 
the others). Using the entropy information, the average EER value 
is increased from $10.01\%$ (for those whom the entropy of their passwords 
is more than $2.7$) to $16.09\%$ (for the others). The 
average method is used 
to fix both thresholds ($8$ and $2.7$). It would be important in 
the future then to investigate more the way of choosing passwords, which may be considered as 
a quality measure in keystroke dynamics research field. Such quality information would 
be useful during the enrollment process in order to enhance the system overall performance.

\section{Conclusion}
  \label{sec_conclusion}
We have presented a new publicly available dataset for keystroke
dynamics.
This dataset is composed of several users who have a different login and password.
We think it is the most realistic keystroke dynamics dataset which is publicly available.
We have statistically verified that:
(a) presenting EER computed with an individual threshold, gives better result than computing the EER with a global threshold (which explains why a lot of keystroke dynamics studies use this method),
(b) using logins gives better performance than using passwords,
(c) using all features during the fusion improves the performance,
(d) the size and the entropy of the password have an impact on the performance.

Keystroke dynamics is an interesting modality, but, it requires strict
conditions during acquisition to avoid capture of noisy samples. 
This may imply an eduction of the user.
As its performance decreases a lot with time, it is necessary to track
the time variability into account which will be the next work on this dataset.
%\IEEEtriggeratref{7}

\bibliographystyle{ieee}

\end{document}